\documentclass[a4paper,fleqn]{cas-dc}
\usepackage{hyperref}

\usepackage[numbers]{natbib}

\usepackage{amsmath}
\usepackage{bm}

\usepackage{amssymb}
\usepackage{dsfont}

\usepackage{algorithm}
\usepackage{algorithmic}
\usepackage{booktabs}
\usepackage{multirow}
\usepackage{adjustbox}

\newcommand{\etal}{\textit{et al}. }

\usepackage[switch]{lineno}

\def\tsc#1{\csdef{#1}{\textsc{\lowercase{#1}}\xspace}}
\tsc{WGM}
\tsc{QE}
\tsc{EP}
\tsc{PMS}
\tsc{BEC}
\tsc{DE}

\begin{document}\sloppy
\let\WriteBookmarks\relax
\def\floatpagepagefraction{1}
\def\textpagefraction{.001}
\shortauthors{Zhiqun Zhao et~al.}

\title [mode=title]{Structure-Preserving Progressive Low-rank Image Completion for Defending Adversarial Attacks}                      

\author[1]{Zhiqun Zhao} 
\fnmark[1]

\address[1]{Department of Electrical Engineering and Computer Science, University of Missouri, Columbia, MO 65203, USA}

\author[2]{Hengyou Wang}
\address[2]{School of Science, Beijing University of Civil Engineering and Architecture, Beijing 100044, China}

\author[1]{Hao Sun}

\author[1]{Zhihai He}
\cormark[1]
\ead{E-mail address: hezhi@missouri.edu}

\begin{abstract}
Deep neural networks recognize objects by analyzing local image details and summarizing their information along the inference layers to derive the final decision. Because of this, they are prone to adversarial attacks. Small sophisticated noise in the input images can  accumulate along the network inference path and produce wrong decisions at the network output.
On the other hand, human eyes recognize objects based on their global structure and semantic cues, instead of local image textures. Because of this, human eyes can still clearly recognize objects from images which have been heavily damaged by adversarial attacks. This leads to a very interesting approach for defending deep neural networks against adversarial attacks. 
In this work, we propose to develop a structure-preserving progressive low-rank image completion (SPLIC) method to remove unneeded texture details from the input images and shift the bias of deep neural networks towards global object structures and semantic cues. We formulate the problem into a low-rank matrix completion problem with progressively smoothed rank functions to avoid local minimums during the optimization process. Our experimental results demonstrate that the proposed method is able to successfully remove the insignificant local image details while preserving important global object structures. On black-box, gray-box, and white-box attacks, our method outperforms existing defense methods (by up to $12.6\%$) and significantly improves the adversarial robustness of the network.
\end{abstract}



\begin{keywords}
adversarial examples, low-rank matrix completion, smoothed rank function, TV norm
\end{keywords}

\maketitle

\section{Introduction}
\label{sec:introduction}

\begin{figure}
\includegraphics[width=\linewidth]{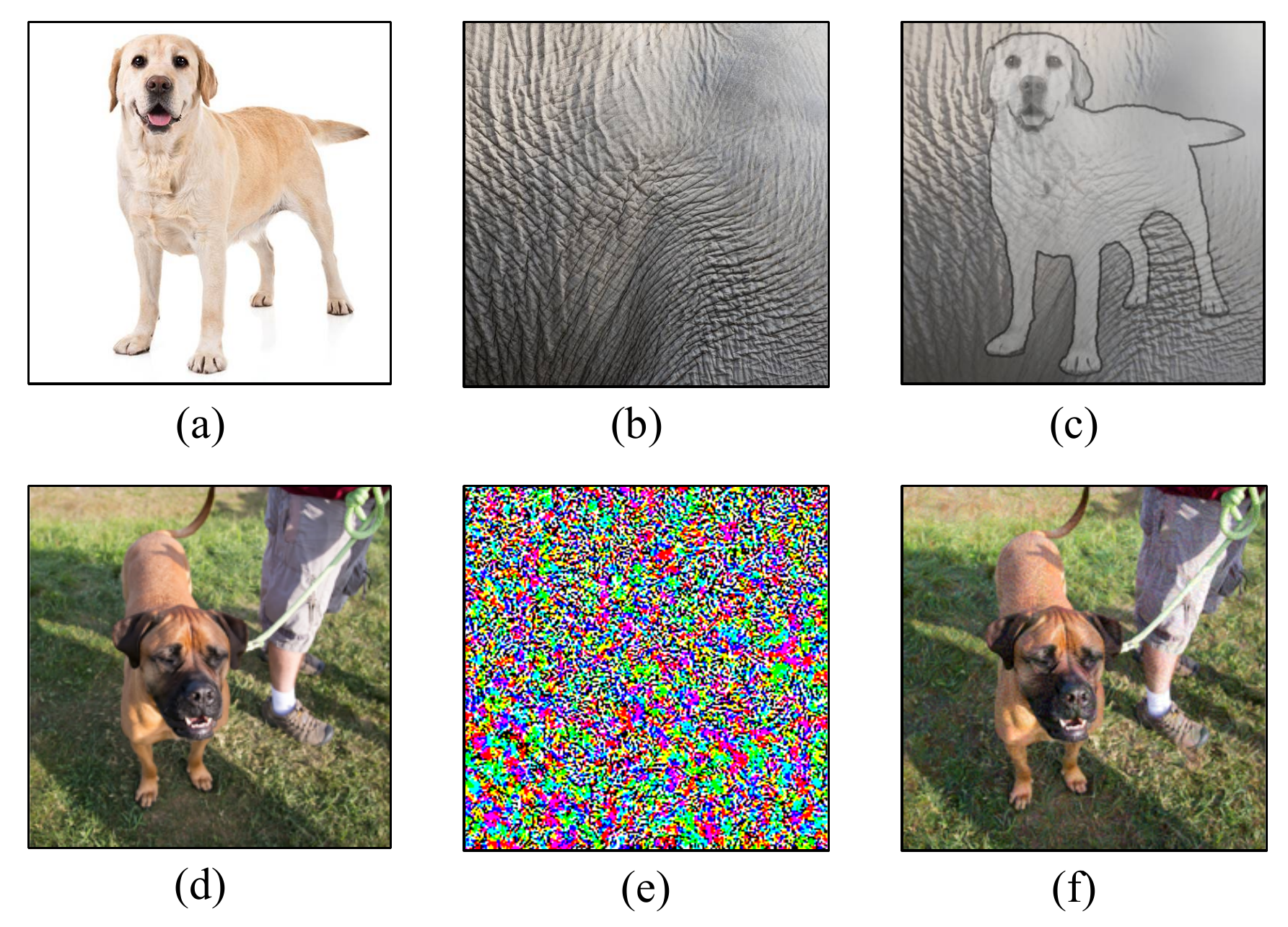}
\caption{(a) a dog image; (b) an India Elephant skin image; (c) the synthesized image of dog from (a) and (b) which is classified as an Indian Elephant \cite{geirhos2018imagenet}; (d) the original dog image; (e) the adversarial attack noise generated by the FGSM white-box attack \cite{goodfellow2015explaining}; and (f) the attacked image being classified as goose.}
\label{fig:example}
\end{figure}

Deep neural networks map the input image pixels into a decision output to classify images, recognize objects, and achieve many other vision analysis tasks. Based on local filtering and pooling, it analyzes pixel values and texture details in each image neighborhood, gradually summarizes the information over the network layers, and produces the final decision at the output layer. Recently, researchers have recognized that deep neural networks are often bias towards image textures instead of semantic structures and global visual cues \cite{geirhos2018imagenet}.
For example, Figure \ref{fig:example} (a) shows an image of dog and (b) shows a texture patch of an Indian elephant. (c) is synthesized from (a) and (b). Deep neural networks, for example, those pre-trained on ImageNet, will often mis-classify image (c) as an Indian elephant. However, human eyes can easily tell that it is still a dog. This suggests that deep neural networks often build their final decision largely upon local image textures, instead of the global object structures, for example, shapes. 

Figure \ref{fig:example} shows another classic example in adversarial attacks \cite{goodfellow2015explaining}: (d) is the original dog image and (e) is the adversarial noise pattern generated by the attack method \cite{goodfellow2015explaining}. The maximum change to each pixel is controlled under $0.7\%$ of the pixel value range. If we add this very small noise onto the original dog image, the deep neural network will classify the result image (f) as goose which is a totally different animal. However, our human eyes have no problem at all in recognizing (f) as a dog since the overall semantic structures are the same.  
This is because, the adversarial noise is uniquely designed so that the error will accumulate along the network inference path, exceed the final decision threshold, and produce a wrong output. 

This example suggests that there is a significant semantic gap between deep neural networks and human visual systems; the emphasis on and bias towards local image texture details cause the network to be prone to adversarial attacks. In the meantime, it also suggests a very interesting approach for defending deep neural networks against adversarial attacks: making the network focus more on semantic structures and global visual cues, instead of local details since the adversarial attacks operate on local pixels and modify their detailed values.

To implement this idea, one possible approach is to develop a structure-preserving image smoothing method to pre-process the image. At the training side, these images are added to the training set to improve the network's capability in capturing global structures. At the test side, this structure-preserving image smoothing will largely remove the adversarial noise hidden in local image textures. Coupled with the structure-oriented training, this method will be able to successfully defend deep neural networks against adversarial attacks.

In this work, we propose to develop a structure-preserving progressive low-rank image completion (SPLIC) method to remove unneeded texture details from the input images and let the deep neural network focus more on global object structures and semantic cues. We formulate the problem into a low-rank matrix completion problem with progressively smoothed rank functions to avoid local minimums during the optimization process. We include total variation constraint to further enhance the capability of our method to capture object structures. Our experimental results demonstrate that the proposed method is able to successfully remove the insignificant local image details and let the network learning focus on global object structures during the learning process. On black-box, gray-box, and white-box attacks, our method outperforms existing defense methods and significantly improve the adversarial robustness of the network.

\section{Related Work and Major Contributions}
\label{sec:related}
In this section, we review related works on adversarial attacks and existing defense methods.
reDuring the past a few years, a number of methods have been developed for defending deep neural networks against adversarial attacks. They can be categorized into two major groups: \textit{adversarial learning} and \textit{transform-based}. 

Adversarial learning \cite{goodfellow2015explaining} aims to improve the neural network robustness. PGD attack is suggested for adversarial training in \cite{madry2018towards} due to its strong perturbation. Kannan \etal \cite{kannan2018adversarial} minimize the logits distance between clean images and adversarial images by introducing the logits pair regularization. Tram{\`e}r \etal \cite{tramer2018ensemble} propose an ensemble adversarial training to resist all attacks, and it is especially effective in black-box defense. Feature scattering \cite{zhang2019defense} uses an unsupervised method to scatter the sample feature in the latent space, and it shifts the previous focus on the decision boundary to the inter-sample structure. A metric learning method has been developed in \cite{mao2019metric} for adversarial training, which utilizes triplet loss to increase the distance between adversarial samples and clean samples.

Transform-based methods aim to process the input images to eliminate adversarial attacks without adding the adversarial images to the training set. 
In \cite{guo2018countering}, five image transformations are applied to remove adversarial perturbations, including image cropping and rescaling, bit-depth reduction, JPEG compression, total variance minimization and image quilting. Ensemble image transforms are used in \cite{taran2019defending,raff2019barrage} to further improve the adversarial defense. A sparse transformation layer (STL) is used in \cite{sun2019adversarial} to project images to the quasi-natural image space. Deep generative models have been used to restore clean images from adversarial images as well, such as GANs \cite{samangouei2018defense} and PixelCNNs \cite{song2018pixeldefend}. A matrix completion method has been developed in \cite{yang2019me} to extract global image structure for network training and testing.
Liu \etal \cite{liu2019feature} propose the JPEG-based defensive compression framework to rectify adversarial examples, called feature distillation. Feature de-noising \cite{xie2019feature} is introduced for defending white-box attacks, which can be considered as a feature transformation method.

\begin{figure*}
\includegraphics[width=1.\linewidth]{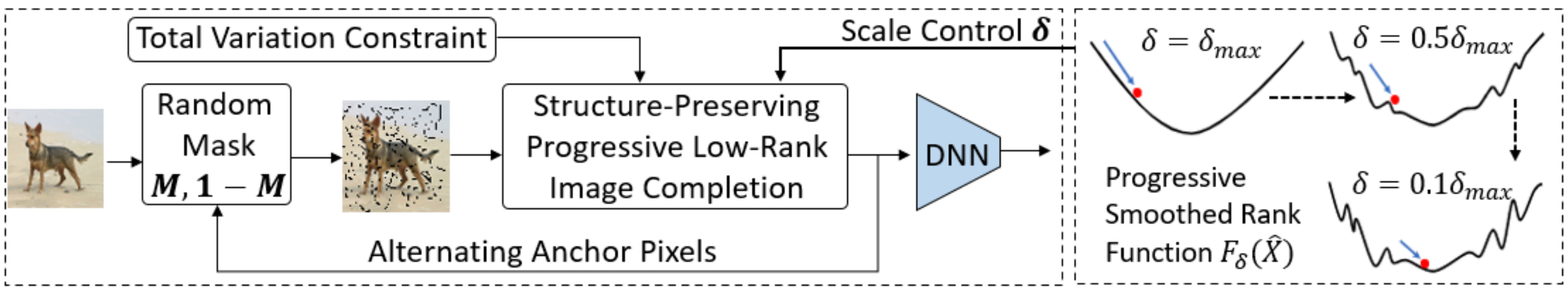}
\caption{Illustration of the proposed structure-preserving low-rank image completion method for defending against adversarial attacks.}
\label{fig:framework}
\end{figure*}

\textbf{Uniqueness of this work and major contributions.} 
Our SPLIC method is image transform-based and it can combine with adversarial training. But, it does not need to change the network architecture. Compared to existing transform-based method, our method is unique since it aims to reduce the bias of deep neural networks towards image texture details where the adversarial attack noise resides. It encourages the network to focus on semantic structures and global visual cues so that it can be more robust under adversarial attacks, just like the human visual system. Comparing to the ME-Net method \cite{yang2019me}, which is closely related to this work, our method is unique in the following aspects: (1) we observe  that the direct approximation of the matrix rank using nuclear norm in the  ME-Net cannot preserve the semantic structures of objects. We have addressed this important issue by introducing the progressively smoothed rank function method. (2) The ME-Net method uses a set of random masks for image completion and generates the final output based on their average. We observe that this averaging method will damage the original image content, especially those structure regions. We have addressed this issue by developing an alternating method for structure-preserving image completion. 

Major contributions of this work can be summarized as follows: (1) We formulate the problem of defending adversarial attacks as a structure-preserving image smoothing problem to remove adversarial noise in local image textures while preserving semantic structures. (2) We incorporate the multi-scale total variation constraint into the smoothed rank function analysis to achieve progressive image smoothing without being trapped into local minimums while maintaining preserving multi-scale object structures. (3) To ensure robust defense against adversarial attacks, we develop an alternated defense method which processes one subset of pixels at each step while using the rest pixels as anchor points. (4) Our extensive experimental results and ablation studies demonstrate that our proposed method outperform existing methods in various defense settings.

\section{Methods} 
\label{sec:high-sensitivity points}
As illustrated in Figure \ref{fig:framework}, we formulate this problem as a structure-preserving low-rank image completion problem. For each input image $\bm{X}$ in both training and test sets, we generate a random mask $\bm{M}$ to select half of the pixels as anchor pixels and the rest as target pixels. We then develop a structure-preserving low-rank image completion (SPLIC) method with total variation constraint to complete the image at target pixel locations using the anchor pixels as constraints. To avoid being trapped into local minimums during rank minimization, we use the method of smoothed rank functions with progressive scale control $\delta$ which can adapt to local object structure at different scales. After the target pixels are completed or smoothed with adversarial noise being largely removed, we alternate the anchor and target pixels, and then apply the above SPLIC method remove the adversarial noise at anchor pixels. 
This SPLIC pre-processing step is applied to all training images and each test image. 
We expect that the learned deep neural network will focus more on semantic structures instead of local texture details, improving its robustness to adversarial attack noise. In the following section, we explain our proposed SPLIC method in more detail.

\subsection{Low-Rank Image Completion Based on Nuclear Norm Minimization}
\label{sec:method:nuclear}

Semantic structures of objects and images are inherently low rank \cite{Wang2018Reweighted}. Recently, methods for low-rank matrix approximation have been developed to characterize the low-rank structures in images \cite{guo2015efficient, liu2017image, wang2017fast, zheng2019exemplar, wang2019multi}. In this paper, we propose to formulate the problem of removing adversarial noise from attacked images while preserving important semantic structure information for successful recognition as a low-rank matrix completion problem. 
Specifically, let $\bm{X}=[{x}_{ij}]_{m\times n} \in \mathbb{R}^{m \times n}$ be the original image of size $m\times n$. $\bm{M} = [m_{ij}]_{m\times n}$ is the random binary mask. If $m_{ij} = 1$, the corresponding image pixel $x_{ij}$ is chosen as the anchor pixel. Otherwise, it is considered as a target pixel. We denoted the set of anchor pixels by $\Omega$.
During low-rank image completion, we attempt to estimate and revise the values of target pixels with the fixed anchor pixels as constraints so that the rank of the recovered image $\hat{\bm{X}} = [\hat{x}_{ij}]_{m\times n}$ is minimized. Specifically, the problem is formulated as 
\begin{equation}
\label{eq:rank-function}
\begin{split}
   \min & \ \mbox{rank}(\hat{\bm{X}}), \\ 
   s.t. & \ x_{ij}=\hat{x}_{ij},\ (i,j)\in \Omega.
\end{split}
\end{equation}
It should be noted that the rank as a function of the matrix is a highly nonlinear and non-convex function \cite{Wang2018Reweighted}, which poses significant challenges for obtaining efficient solutions for the problem in (\ref{eq:rank-function}). More importantly, the solution is often trapped into local minimums. 
To address this issue, the nuclear norm $||\hat{\bm{X}}||_*$ of matrix $\bm{X}$ is often used to approximate the rank of matrices, which leads to
a convex minimization problem with highly efficient solutions available. 
Let $\{\sigma_k(\bm{\hat{X}})\}$, $1\le k\le l$, $l=\min(m, n)$, be the set of singular values of matrix $\bm{\hat{X}}$.
Then, the rank of $\bm{\hat{X}}$ is the number of non-zero entries in $\{\sigma_k(\bm{\hat{X}})\}$. However, the nuclear norm is the summation of all singular values. The nuclear norm approximation of the optimization problem in (\ref{eq:rank-function}) is given by 
\begin{equation}
\label{eq:nuclear-norm}
\begin{split}
   \min & \ ||\hat{\bm{X}}||_* = \sum_{k=1}^K \sigma_k(\bm{\hat{X}}), \\ 
   s.t. & \ x_{ij}=\hat{x}_{ij},\ (i,j)\in \Omega.
\end{split}
\end{equation}

\subsection{Progressive Smoothed Rank Functions with Total Variation Constraint}
\label{sec:method:splic}

In this work, we have found that the nuclear norm does not provide an effective approximation of the original rank function, especially for images or objects with complex semantic structures at different spatial scales. We observe that, geometrically, smooth terms generally lie much closer to the essential rank function than nuclear norm. In the meantime, we wish to take advantage of the convex nature of the nuclear norm so that the optimization process will not be trapped into local minimums.
To address this issue, we propose to use the smoothed rank function method developed in \cite{ghasemi2011srf} to better preserve the important structure information.  
Given a matrix $\hat{\bm{X}}$, its smoothed rank function is defined based on Gaussian smoothing of its singular values:
\begin{equation}
\label{eq:smoothed-gaussion}
\begin{aligned}
  F_{\delta}(\hat{\bm{X}})=l-\sum_{k=1}^l e^{-\frac{\sigma_{k}^2(\hat{\bm{X}})}{2\delta^2}}
\end{aligned}
\end{equation}
This smoothed rank function well approximates the original rank of the matrix, when $\delta$ approaches 0.
For example, considering a matrix $\hat{\bm{X}}$ with a rank of $k_0$. The first $k_0$ singular values are positive, $\sigma_k(\hat{\bm{X}}) > 0$ for $k\le k_0$. The rest singular values are zeros, $\sigma_k(\hat{\bm{X}}) = 0$ for $k_0< k\le l$. 
In this case, when $\delta\rightarrow 0$, we have 
\begin{equation}
    e^{-\frac{\sigma_{k}^2(\hat{\bm{X}})}{2\delta^2}}  
    =\left\{ 
    \begin{array}{ll}
        0, & \quad 0\le k\le k_0,\\
        1, & \quad  k_0< k\le l, 
    \end{array}
    \right.
\end{equation}
Therefore, according to (\ref{eq:smoothed-gaussion}), $F_{\delta}(\hat{\bm{X}}) = k_0$.
Figure \ref{fig:framework} (right) shows the smoothed rank function with progressive control $\delta$. When $\delta$ is large, it is a convex function. Based on this rank function, the algorithm can guide the optimization towards the region of global minimum. Then, the method gradually reduces the value of $\delta$ and refines the scale of gradient search. This progressive optimization can successfully avoid the local minimum while enjoying the advantage of local convex optimization. 
This gradual tuning technique for minimizing non-convex functions is referred to as graduated non-convexity \cite{ghasemi2011srf}.

In this work, we observe that the smoothed rank function can obtain better  performance in matrix completion than nuclear norm which was used in the ME-Net method \cite{yang2019me}. However, it still suffers from performance degradation when the image has high intrinsic rank structures or has noise density. It is not able to efficiently remove sparse adversarial noise with high density due to the absence of an proper regularization scheme. Furthermore, they cannot effectively maintain the smoothness of local neighborhood pixels in the smooth regions affected by adversarial noise. To address this issue, we propose to incorporate the total variation (TV) constraint \cite{lu2014smoothed} into the progressive smoothed rank optimization problem.
Mathematically, our  SPLIC optimization problem can be formulated as:
\begin{equation}
\label{eq:problem}
\begin{aligned}
  & \min_{\hat{\bm{X}}} F_{\delta}(\hat{\bm{X}}) + \lambda\cdot C(\hat{\bm{X}})  \\ 
  & s.t. \ x_{ij}=\hat{x}_{ij},\ (i,j)\in \Omega.
\end{aligned}
\end{equation}
where $\lambda$ is a weighting parameter which will be analyzed in our ablation studies. 
$C(\hat{\bm{X}})$ is the TV constraint. 
Since the original TV-norm is hard to compute the gradient directly, we rewrite the TV constraint function as follows:
\begin{equation}
\label{formula:tv-norm}
\begin{aligned}
 C(\hat{\bm{X}}) \! = \! & \sum_{i \! = \! 1}^{m \!- \!1} \! \sum_{j \! = \! 1}^{n \! - \! 1} \frac{(\hat{x}_{i,j} \! - \! \hat{x}_{i \! + \! 1, j})^2 \! + \! (\hat{x}_{i,j} \! - \! \hat{x}_{i, j \! + \! 1})^2}{2} \\
  & \! + \! \sum_{i \! = \! 1}^{m \! - \! 1} \! \frac{(\hat{x}_{i,n} \! -\! \hat{x}_{i \! + \! 1,n})^2}{2} \! + \! \sum_{j \! = \! 1}^{n \! - \! 1} \! \frac{(\hat{x}_{m,j} \! - \! \hat{x}_{m,j \! + \! 1})^2}{2},
\end{aligned}
\end{equation}
where the first entry in the summation are variations for pixels inside the image and the last two entries are for pixels on the horizontal and vertical edges.

\subsection{Solution to the SPLIC Optimization Problem}
\label{sec:method:splic-sol}

In this section, we derive a gradient descent-based numerical solution for the SPLIC optimization problem in (\ref{eq:problem}). To this end, we need to determine the derivatives of $F_{\delta}(\hat{\bm{X}})$
and $C(\hat{\bm{X}})$ with respect to $\hat{\bm{X}}$.

We first introduce the definition of \textit{\textbf{absolutely symmetric function}} \cite{lewis1995convex}. Given a vector $\bm{\gamma}$ in $\mathbb{R}^q$, we sort its vector elements in a non-increasing order to form a new vector $\hat{\bm{\gamma}}$. A function  $f:\mathbb{R}^q \rightarrow \mathbb{R}$ is absolutely symmetric if $f(\gamma)=f(\hat{\gamma})$ for any vector $\gamma$ in $\mathbb{R}^q$. For matrix $\hat{\bm{X}}$, its singular value decomposition (SVD) is 
\begin{equation}
\hat{\bm{X}} = \bm{U} \cdot\mbox{diag} \{ \sigma_1(\hat{\bm{X}}), \cdots,\sigma_l(\hat{\bm{X}}) \}\cdot  \bm{V}^T,    
\end{equation}
where $\bm{U}$ and $\bm{V}$ are right and left singular vector matrices. 
We can see that the following function
\begin{equation}
    F_{\delta}(\bm{z}) = l-\sum_{k=1}^l e^{-\frac{z_l^2}{2\delta^2}}
\end{equation}
is absolutely symmetric. 
According to the Theorem 3.1 of \cite{lewis1995convex} and \cite{ghasemi2011srf},  the sub-gradient of $F_{\delta}(\hat{\bm{X}})$ can be calculated as follows:
\begin{equation}
\label{formula:gaussion-gradient}
\begin{aligned}
  \nabla F_{\delta}(\hat{\bm{X}}) = \bm{U} \cdot \mbox{diag}\{\frac{\sigma_{1}}{\delta^2}e^{-\frac{\sigma_{1}^2}{2\delta^2}},\cdots,\frac{\sigma_l}{\delta^2}e^{-\frac{\sigma_{l}^2}{2\delta^2}}\} \cdot \bm{V}^T.
\end{aligned}
\end{equation}
For the total variation term $C(\hat{\bm{X}})$ in equation (\ref{formula:tv-norm}), its derivative with respect to 
$\hat{\bm{X}}$ is given by 
\begin{equation}\label{formula:tv-gradient}
\begin{aligned}
  \nabla C(\hat{\bm{X}}) \! = \!
  \begin{cases}
   2\hat{x}_{i,j} \! - \! \hat{x}_{i + 1,j} \! - \! \hat{x}_{i,j + 1}, \! & \! \mbox{inside pixels} \\
    \hat{x}_{i,j} \! - \! \hat{x}_{i,j + 1}, \! & \! i \! = \! m \\
    \hat{x}_{i,j} \! - \! \hat{x}_{i + 1,j}, \! & \! j \! = \! n 
    \end{cases}
\end{aligned}
\end{equation}
With the gradient of $F_{\delta}(\hat{\bm{X}})$ and the gradient of $C(\hat{\bm{X}})$ being obtained by equations (\ref{formula:gaussion-gradient}) and (\ref{formula:tv-gradient}), we are ready to use gradient descent algorithm to solve the problem (\ref{eq:problem}). 

\subsection{Alternated SPLIC and Algorithm Summary}
\begin{figure}
\centering
{\includegraphics[width=1\linewidth]{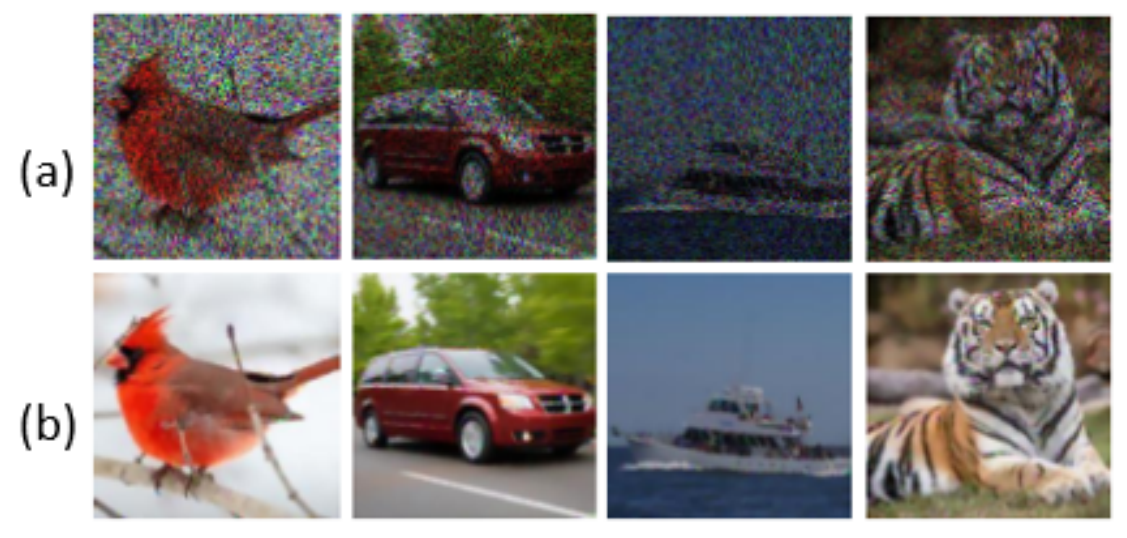}}
\caption{Example results by SPLIC: (a) the noise images and (b) the final SPLIC results.}
\label{fig:reconstructed-example}
\end{figure}

In our SPLIC method, we randomly select $50\%$ of pixels as anchor points. At these anchor points, the pixel values are fixed as constraints in the optimization problem (\ref{eq:problem}). We observe that these constraints are very important for the robustness of our SPLIC method to avoid algorithm divergence. Once the rest $50\%$ pixels have been re-estimated by our SPLIC method, we will alternate the SPLIC process, using them as the anchor points, and re-estimate the values of the original anchor points. 
The SPLIC algorithm is summarized in  Algorithm \ref{alg1}.
Figure \ref{fig:reconstructed-example} shows four example results by our SPLIC method. The first row shows the noise images heavily damaged by the adversarial attacks. The second row shows the recovered images by our SPLIC method. We can see that it is able to remove the adversarial noise, as well as the detailed textures of the original image content, while largely maintaining the semantic structure of the objects, which are very important for image recognition and machine learning. During network training, we first use the SPLIC method to pre-process all training images before passing them to the network. During testing, we first apply the SPLIC to process the input image which might have been corrupted by the adversarial attack. Then, we pass the image to the target network for analysis. This will reduce the bias of the network towards detailed image textures where the adversarial noise hides and encourage the learned network focus on global semantic structure and visual cues, just like the human visual system. As a result, it will significantly improve the robustness of the network.

\begin{algorithm}
\caption{structure-preserving progressive low rank image completion (SPLIC)}
\label{alg1}
\begin{algorithmic}[1]
\REQUIRE $\bm{X} \in \mathbb{R}^{m \times n}$, $\bm{M} \in \mathbb{R}^{m \times n}$, pre-defined rank $r \leq l$, criteria $\varepsilon$ and the maximum iteration number $\mbox{maxiter}$.
\STATE \textbf{Initialize:} $\hat{\bm{X}}_t = \bm{X} \odot \bm{M}$, $\lambda=0.02$, $\rho=0.45$, $\mu=0.5$, $t=0$, and $\delta$ is set as the largest singular value of $\hat{\bm{X}}_{t}$.
\WHILE{$\frac{\|\hat{\bm{X}}_{t+1}-\hat{\bm{X}}_t \|_{F}}{mn}>\varepsilon$ and $t< \mbox{maxiter}$}
\FOR{$i=1$ to $7$} 
\STATE Compute SVD of $\hat{\bm{X}}_t$; \\ 
$\hat{\bm{X}}_t = \bm{U}\bm{S}\bm{V}^T, \bm{S} = \mbox{diag}\{\sigma_{1},\cdots,\sigma_{l}\}$;
\STATE Set $\sigma_{r+1},\cdots,\sigma_{l}$ to zeros, \\
$\bm{S} = \mbox{diag}\{\sigma_{1},\cdots,\sigma_{r}, 0, \cdots, 0\}$;
\STATE Compute the gradient of $F_\delta(\hat{\bm{X}}_t)$ by equation (\ref{formula:gaussion-gradient});
\STATE Compute the gradient of $C(\hat{\bm{X}})$ by equation (\ref{formula:tv-gradient});
\STATE Update $\widetilde{\bm{X}}_{t+1}$: \\
$\widetilde{\bm{X}}_{t+1} = \hat{\bm{X}}_t-\mu(\nabla F_{\delta}(\hat{\bm{X}}_t) + \lambda \nabla C(\hat{\bm{X}}_t))$;
\STATE Compute the projection: \\ $\hat{\bm{X}}_{t+1} = (\bm{1-M}) \odot \widetilde{\bm{X}}_{t+1} + \bm{M} \odot \bm{X}$;
\STATE $t = t + 1$;
\ENDFOR
\STATE Update the smoothness parameter: $\delta=\rho\delta$;
\ENDWHILE
\ENSURE $\hat{\bm{X}} = \hat{\bm{X}}$.
\end{algorithmic}
\end{algorithm}

In this section, we follow the procedures in existing papers to evaluate the performance of our SPLIC method and compare its performance with the state-of-the-art methods. 

\subsection{Experimental Settings}
\label{sec:experiments:setting}

We evaluate all defense performance in the following three attack scenarios: (1) white box attackers where the attacker has full knowledge about the network and the defense method, (2) black-box attackers where the attacker has no knowledge about the network and the defense network, and (3) gray-box attackers where the attacker knows the network but does not know the defense method. 
In our experiments, we conduct performance comparison with existing papers on four attack methods, the FGSM, PGD and BPDA, and CW \cite{carlini2017towards} methods, as reviewed in the Related Work section. 
We use the publicly available package FoolBox \cite{rauber2017foolbox} for implementation of these attackers.
Following prior papers, we conduct performance comparisons on two benchmark datasets, the  \textbf{CIFAR-10} \cite{krizhevsky2009learning} and the \textbf{SVHN} datasets \cite{netzer2011reading}. 

All the algorithms run on a desktop computer with an Intel core i7-7800X 3.50 GHz CPU, one Nvidia GTX 1080 Ti GPU, 64 GB of RAM, and Ubuntu 18.04.
The perturbation in all adversarial attacks are constrained within an $\epsilon$-ball based on the $L_{\infty}$ distance and we set $\epsilon = 8/255$. In iterative attack methods of PGD and BPDA, we set the single step size to $2/255$ and set the number of iterations to $7$. In the CW attack, we fix the confidence level $\kappa=20$ and the binary search step size as $5$. The learning rate and the number of iterations are set as $0.005$ and $1000$ respectively.

\subsection{Convergence Analysis of the SPLIC  Algorithm}
\label{sec:experiments:splic-convergence}

\begin{figure}
\centering
\includegraphics[width=1.0\linewidth]{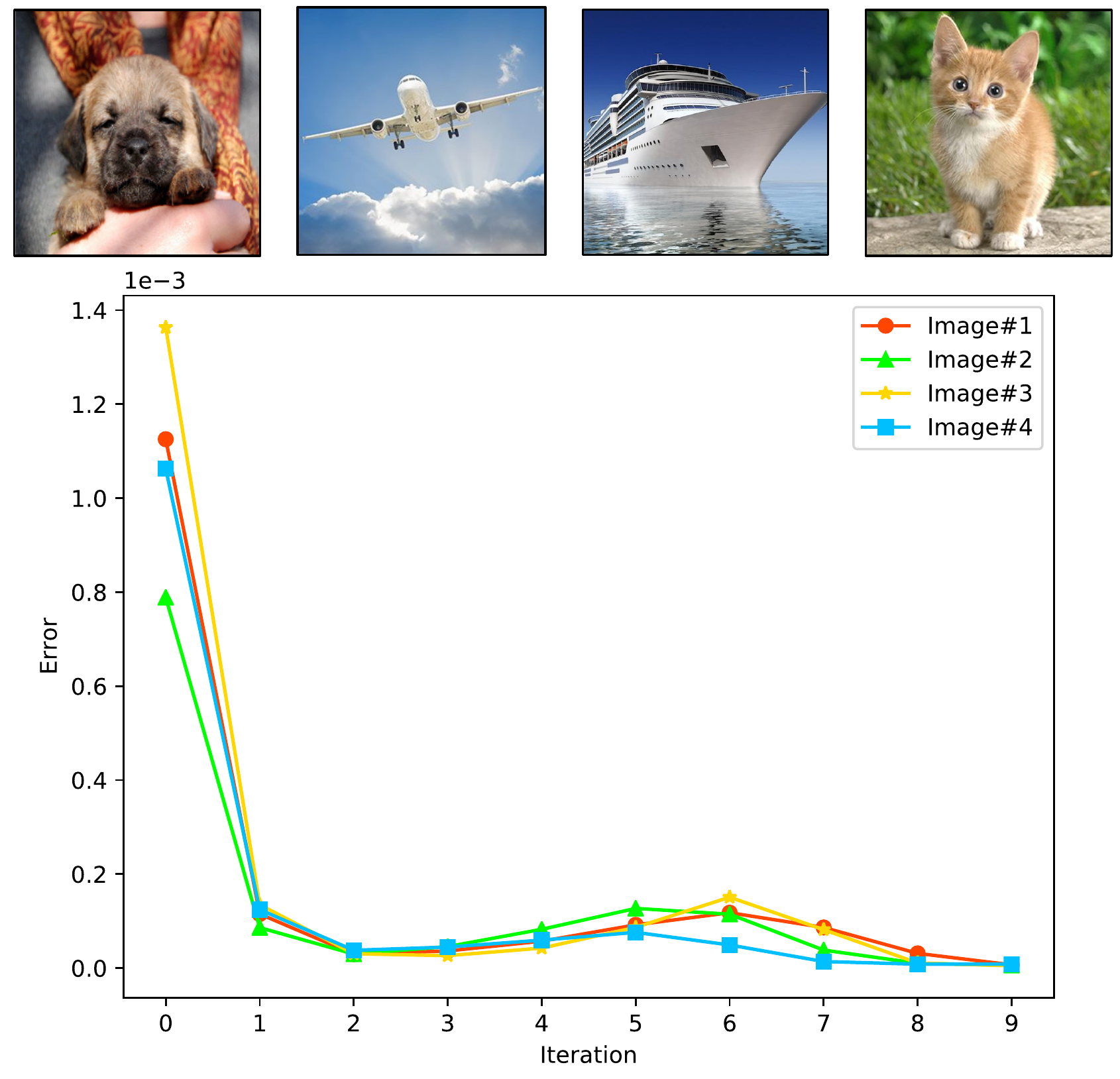}
\caption{Convergence behavior of the proposed SPLIC algorithm.}
\label{fig:convergence}
\end{figure}

In our proposed SPLIC method, we gradually reduce the value of $\delta$ to control the smoothness of the rank function. When the $\delta$ approaches zero, the objective function $F_{\delta}(\hat{\bm{X}})$ will approach to the real rank function. But decreasing $\delta$ to zero will result in a highly non-smoothed $F_{\delta}(\cdot)$, and the gradient projection method might be trapped in local minimums and fail to converge.  For a given $\delta$, this technique uses the minimizer in the previous iteration (i.e. the previous larger $\delta$) as the new starting point to search for the minimum solution in the current iteration (i.e. the current $\delta$). Figure \ref{fig:convergence} shows the convergence behaviors of four sample images. We can see that the relative errors decrease significantly and reach a stable solution after several iterations. We also find that the experimental convergence results are consistent to the above analysis.

\subsection{SPLIC with Different Ranks}
\label{sec:sup:rank}

\begin{figure}
\centering
\includegraphics[width=1\linewidth]{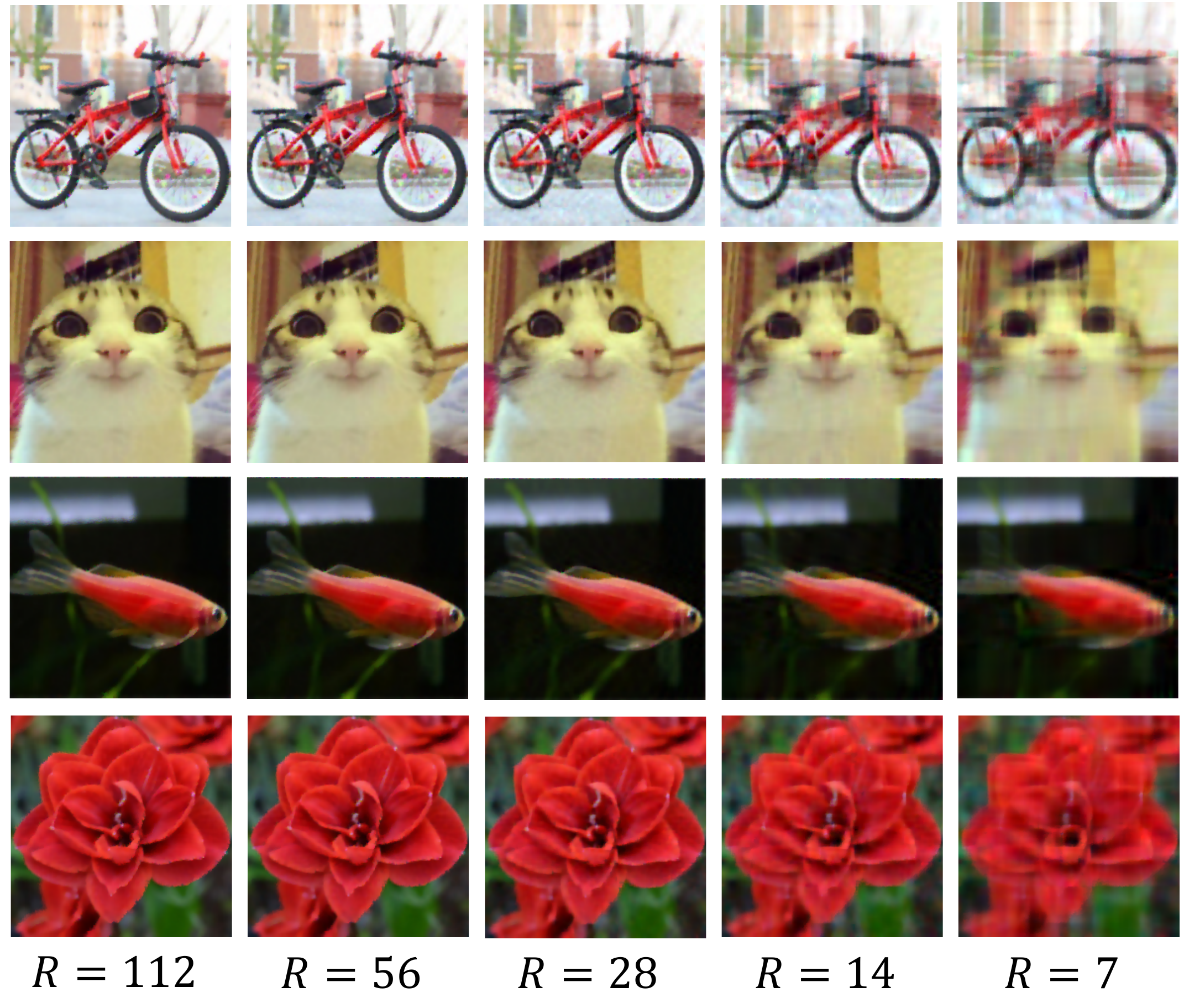}
\caption{SPLIC images with different ranks at $r=112$, $r=56$, $r=28$, $r=14$ and $r=7$.}
\label{fig:rank-samples}
\end{figure}

In this work we consider the image as a matrix. For example, an image of size $112\times 112$ is considered as a matrix of dimension $112\times 112$. Its full rank should be 112. If we reduce the rank of the matrix using the SPLIC method, the image will lose more details. Figure \ref{fig:rank-samples} show example images being reduced to different ranks, such as 56, 28, 14, and 7. We can see that the images details are reduced at lower ranks, but the overall object structure is largely maintained. 

\begin{figure}
\centering
\includegraphics[width=1\linewidth]{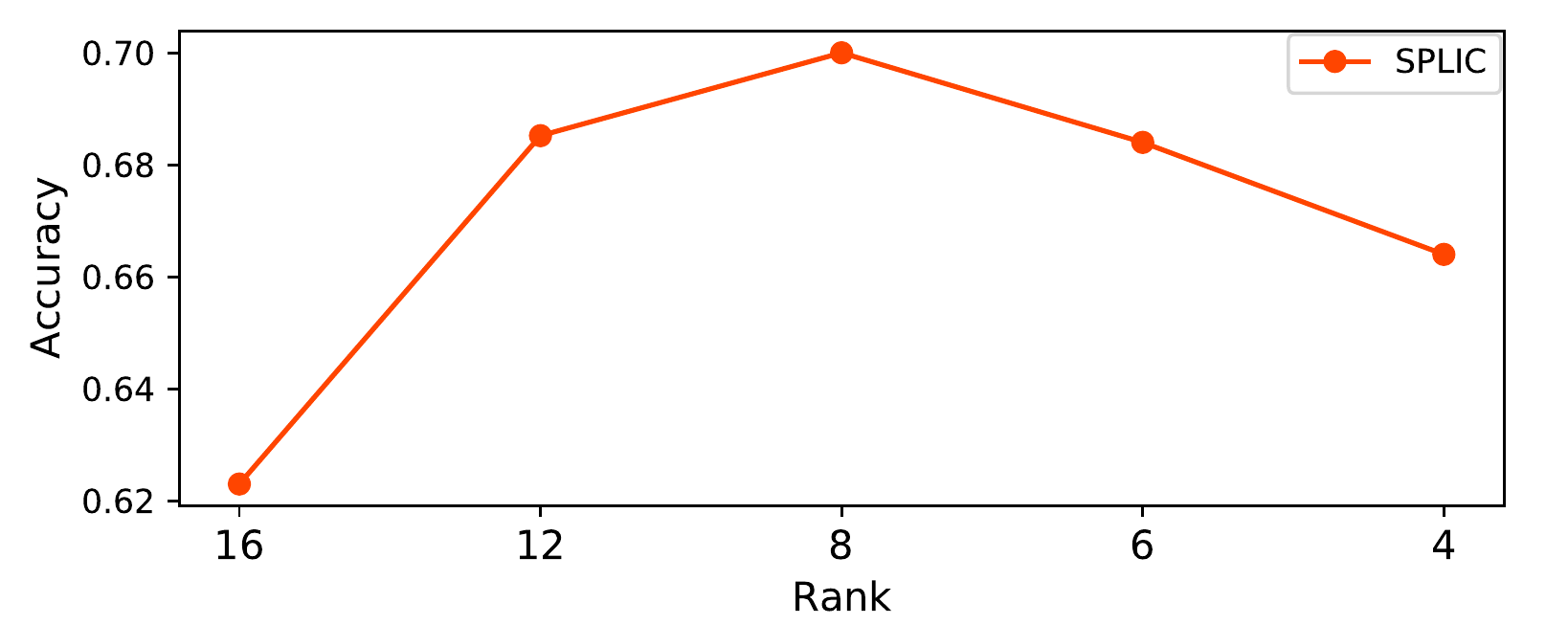}
\caption{SPLIC accuracy with different ranks.}
\label{fig:rank-acc}
\end{figure}

In the following, we apply the rank reduction to the CIFAR-10 images and reduce their ranks to a target number. 
Figure \ref{fig:rank-acc} shows the final classification accuracy of the images for different target ranks. 
We can see that, when we reduce the rank from 32 to 16, 12, and 8, the classification accuracy is increasing. This is because more noise has been removed. Here, the image size is 32, which is also the full rank of the matrix. But, if we keep decrease the target rank, the final classification accuracy will decrease, because more original image content has been damaged by the rank reduction process. Thus, in our experiments,  we set the target rank to $r=R/4$ (i.e. $r=8$).

Figure \ref{fig:ori-adv-srf-splic-samples} shows several examples of reconstructed images from the CIFAR-10 and SVHN datasets. The first row shows the original images. The second row shows the images with adversarial attacks. The third row shows the images reconstructed by our SPLIC method. We can see that the SPLIC method is able to successfully remove the adversarial noise while largely maintain the semantic structures.

\subsection{Performance Comparison with Existing Methods}
\label{sec:experiments:comparison}
In the following, we evaluate  our SPLIC methods under three different attack scenarios and compare its performance with existing methods.

\subsubsection{Defense against black-box attackers}
Table \ref{tab:black-box-defense} summarizes the defense performance of our SPLIC methods under three different black-box attacks: FGSM, PGD, and CW on the CIFAR-10 and SVHN datasets, and performance comparisons with existing methods. 
We also include the results on the clean images without any attacks. The Vanilla method means no defense is applied. 
We can see that on these black-box attacks, our SPLIC method outperforms the current best method, ME-Net \cite{yang2019me}. 
It should be noted that performance gain is not very significant because the defense accuracy is already very high, very close to the accuracy on the clean images without any attack. 
For example, on the CIFAR-10, the best accuracy on the clean images is $94.9\%$. Under the powerful CW attack, our SPLIC method can achieve the accuracy of $93.6\%$. On the SVHN dataset, this gap is even smaller. We can also see that on the clean images, our method is able to maintain the important semantic structure information for recognition, achieving near the best accuracy. Some methods do not report the results of FGSM and CW, thus we represent them by $-$ in the table.

\begin{table}
\centering
{\small
\caption{Defense performance under black-box attacks.}
\label{tab:black-box-defense}
\begin{tabular}{c|c|c|c|c}
\toprule
\multicolumn{5}{c}{CIFAR-10} \\
\hline

Method  &   Clean  &  FGSM  &  PGD  &  CW  \\
\hline
Vanilla  &  $93.4\%$  & $24.8\%$  &  $7.6\%$  &  $9.3\%$  \\
Madry \cite{madry2018towards}  &  $79.4\%$  &  $67.0\%$  &  $64.2\%$  &  $78.7\%$  \\
Thermometer \cite{Buckman2018Thermometer}  &  $87.5\%$  &  $-$  &  $77.7\%$  &  $-$  \\
TLA-RN \cite{mao2019metric}  &   $81.0\%$  &  $-$  &  $66.0\%$  &  $-$  \\
TLA-SA \cite{mao2019metric}  &   $86.2\%$  &  $-$  &  $61.7\%$  &  $-$  \\
TLA \cite{mao2019metric}  &   $86.2\%$  &  $-$  &  $70.6\%$  &  $-$  \\
ME-Net \cite{yang2019me} &   $\textbf{94.9\%}$  &  $\textbf{92.2\%}$  &  $91.8\%$  &  $\textbf{93.6\%}$  \\
\hline
SPLIC-Net (Ours) &  $94.0\%$  &  $91.0\%$  &  $\textbf{92.2\%}$  &  $\textbf{93.6\%}$  \\
\hline
\hline
\multicolumn{5}{c}{SVHN} \\
\hline
Vanilla  &   $95.0\%$  &  $31.2\%$  &  $8.6\%$  &  $20.4\%$  \\
ME-Net \cite{yang2019me} &   $\textbf{96.0\%}$  &  $91.8\%$  &  $91.1\%$  &  $95.5\%$  \\
\hline
SPLIC (This Work) &  $95.8\%$  &  $\textbf{92.9\%}$  &  $\textbf{93.0\%}$  &  $\textbf{95.7\%}$  \\
\bottomrule
\end{tabular}
}
\end{table}

\subsubsection{Defense against white-box attacks}
Table \ref{tab:white-box-defense} summarizes the performance of our SPLIC method under the white-box BPDA attack on the CIFAR-10 and SVHN datasets. The attack method knows both the network and the defense methods. 
We can see that our SPLIC method outperforms existing  state-of-the-art methods by a large margin. For example, on the CIFAR-10 dataset, it improves the accuracy by $8.7\%$. On the SVHN dataset, the performance gain is $6.1\%$, which is quite significant. We can also see that the white-box BPDA attack is very powerful, making it much more challenging to defend. This is because the white-box attack has full knowledge about the defense method and has the opportunity to learn the behavior of the defense method and re-adjust the adversarial attack noise.

\begin{table}
\centering
\caption{Defense performance under white-box attacks.}
\label{tab:white-box-defense}
\scalebox{1}{
\begin{tabular}{c|c|c}
\toprule
Dataset  &  Method  &  BPDA  \\
\hline
\multirow{7}{*}{CIFAR-10}  &  Vanilla  &  $0.0\%$  \\
&  TV Mnimization \cite{guo2018countering}  &  $14.7\%$  \\
&  TLA-RN \cite{mao2019metric}  &  $52.5\%$  \\
&  TLA-SA \cite{mao2019metric}  &  $53.5\%$  \\
&  TLA \cite{mao2019metric}  &  $53.9\%$  \\
&  ME-Net \cite{yang2019me}  &  $59.8\%$  \\
\cline{2-3}
&  SPLIC (This Work)  &  $\textbf{68.5\%}$  \\
& Gain & \textbf{+8.7}\% \\
\hline
\hline
\multirow{4}{*}{SVHN}  &  Vanilla  &  $0.0\%$  \\
&  Madry \cite{madry2018towards}  &  $52.5\%$  \\
&  ME-Net \cite{yang2019me}  &  $74.7\%$ \\
\cline{2-3}
&  SPLIC (This Work) &  $\textbf{80.8\%}$  \\
& Gain & \textbf{+6.1}\% \\
\bottomrule
\end{tabular}
}
\end{table}

\subsubsection{Defense against gray-box attacks}
Table \ref{tab:gray-box-defense} summarizes the results on the CIFAR-10 and SVHN under gray-box attacks without network training. We compare our method against three defense methods under the gray-box attack scenarios. 
Since the ME-Net \cite{yang2019me} only reported the results after the training on reconstructed images. We used their released code to obtain the results in Table \ref{tab:gray-box-defense}.
We can see that our SPLIC method outperforms existing methods by large marings. On the CIFAR-10 and SVHN datasets, the performance gains under the PGD attack are $12.6\%$ and $9.1\%$, respectively.

\begin{table}
\centering
\caption{Defense performance under gray-box attacks without network training.}
\label{tab:gray-box-defense}
\scalebox{1}{
\begin{tabular}{c|c|c|c}
\toprule
Dataset  &  Method  &  FGSM  &  PGD \\ 
\hline
\multirow{4}{*}{CIFAR-10}  &  USVT  & $12.1\%$  & $12.1\%$ \\ 
&  Soft-Imp  & $30.1\%$  & $27.3\%$  \\ 
&  ME-Net  & $31.2\%$  & $31.3\%$  \\ 
\cline{2-4}
&  SPLIC  &  $\textbf{40.9\%}$  & $\textbf{43.6\%}$ \\ 
& Gain  & \textbf{+9.7\%} & \textbf{+12.6\%} \\ 
\hline
\hline
\multirow{4}{*}{SVHN}  &  USVT  & $36.7\%$  &  $34.2\%$  \\ 
&  Soft-Imp  &  $56.1\%$  &  $55.6\%$  \\ 
&  ME-Net  &  $55.4\%$  &  $54.5\%$  \\ 
\cline{2-4}
&  SPLIC  &  $\textbf{59.8\%}$  &  $\textbf{63.6\%}$  \\ 
& Gain & \textbf{+3.6\%} & \textbf{+9.1\%} \\ 
\bottomrule
\end{tabular}}
\end{table}

\subsubsection{Comparison with existing smoothing methods} 
\begin{figure}
\centering
{\includegraphics[width=1\linewidth, height = 0.40\linewidth]{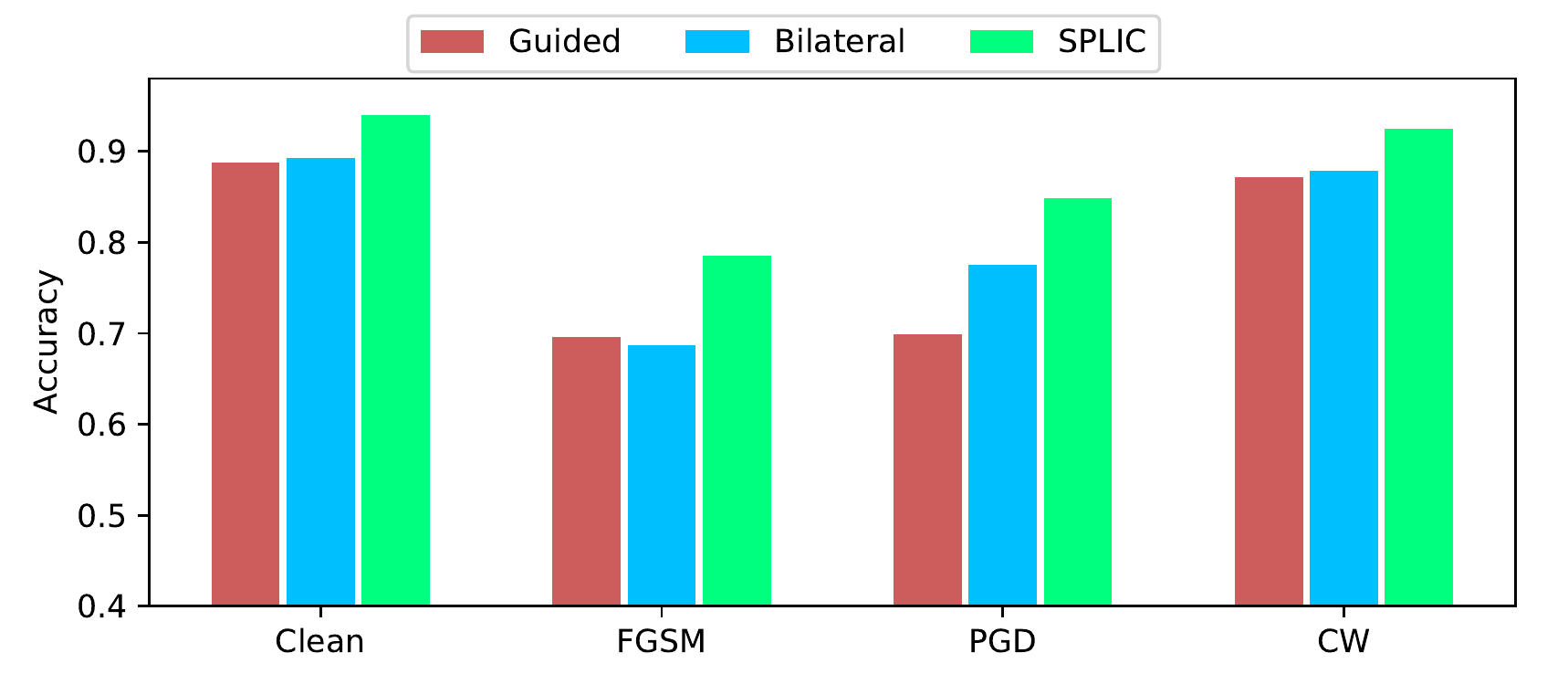}}
\caption{Comparison with existing image smoothing method.}
\label{fig:smooth-compare}
\end{figure}

Structure-preserving image smoothing has been studied in the image de-noising literature \cite{he2015fast}. In this work, we have developed the SPLIC method for structure-preserving image completion. 
In the following experiment, we compare our SPLIC method with two existing image smoothing methods based on bilateral image filtering \cite{tomasi1998bilateral} and  edge-guided image de-noising \cite{he2015fast}.  After processed by these methods, we retrain the network and evaluate the defense performance on the CIFAR-10 under white-box attacks. Figure \ref{fig:smooth-compare} shows their classification accuracy comparison. 
We can see that, under FGSM, PGD and CW attacks, our method outperforms existing image smoothing methods. On the clean images, it can preserve the original image semantic structures much better than other methods. This is because our progressive smoothed rank function with total variation constraint can successfully capture and preserving the multi-scale semantic structures in the image, which are very important for network learning and image recognition.

\begin{figure*}
\centering
\includegraphics[width=1\linewidth]{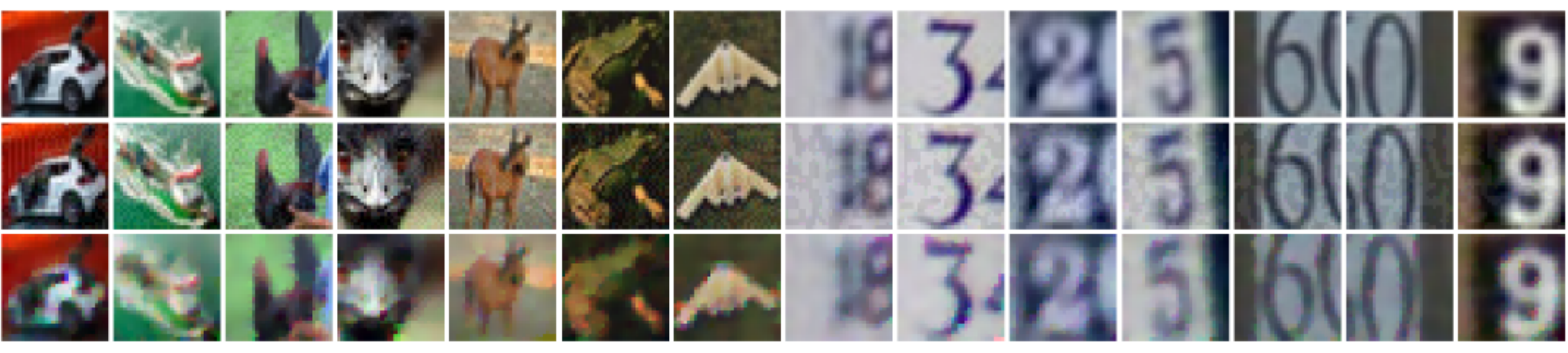}
\caption{Image Samples: the first row are original images, the second row are adversarial images, the third row are the reconstructed images by our  SPLIC method.}
\label{fig:ori-adv-srf-splic-samples}
\end{figure*}

\subsection{Ablation Studies}
In the following, we provide ablation studies to further understand our SPLIC method.

\subsubsection{Impact of the weighting parameter $\lambda$. } 
\begin{figure}
\centering
{\includegraphics[width=1\linewidth]{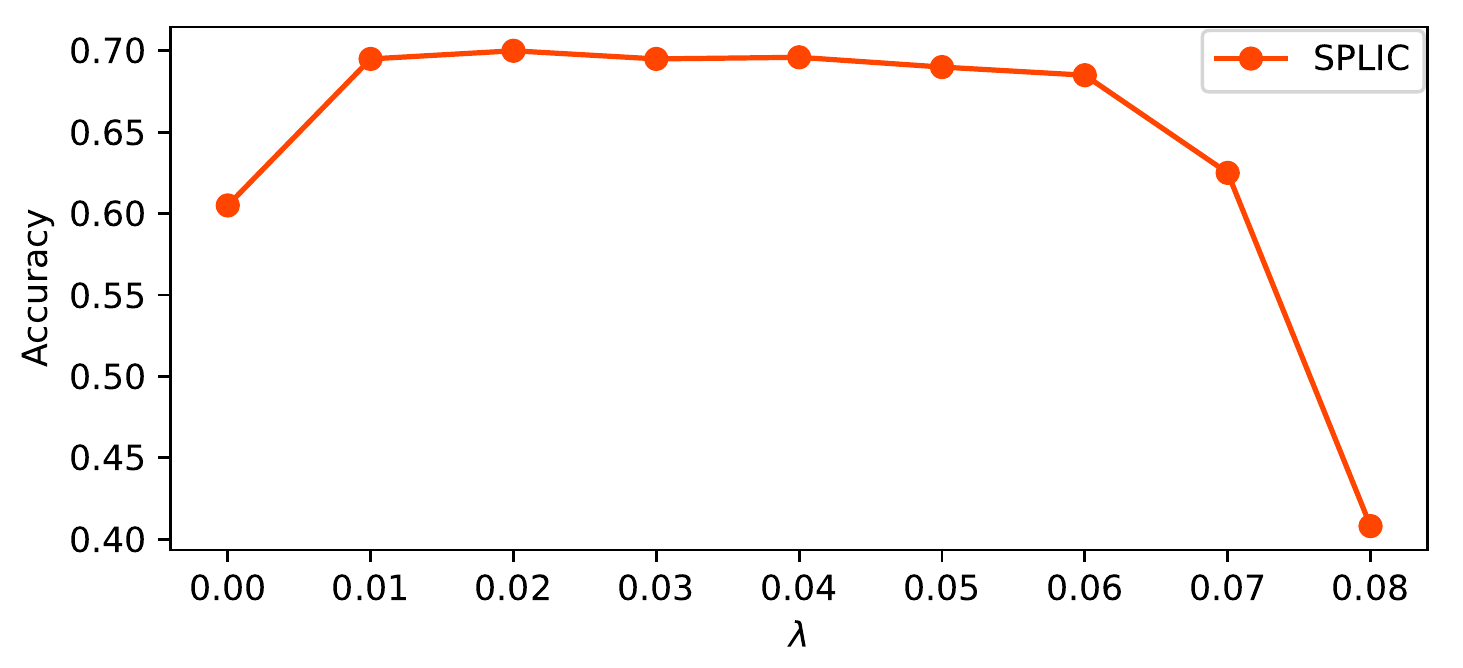}}
\caption{SPLIC accuracy with different $\lambda$.}
\label{fig:splic-params}
\end{figure}

The SPLIC optimization problem in (\ref{eq:problem}) has two objective functions weighted by the control parameter $\lambda$. 
Figure \ref{fig:splic-params} shows the classification accuracy obtained by our SPLIC method with different $\lambda$ on the CIFAR-10 dataset under white-box BPDA attack. We can see that the best performance is achieved for $\lambda$ within the range of $[0.01, 0.05]$. In our experiment, we set $\lambda$ to be $0.02$.

\subsubsection{Visualization of image samples in the learned feature space}
\begin{figure}
\centering
\includegraphics[width=\linewidth]{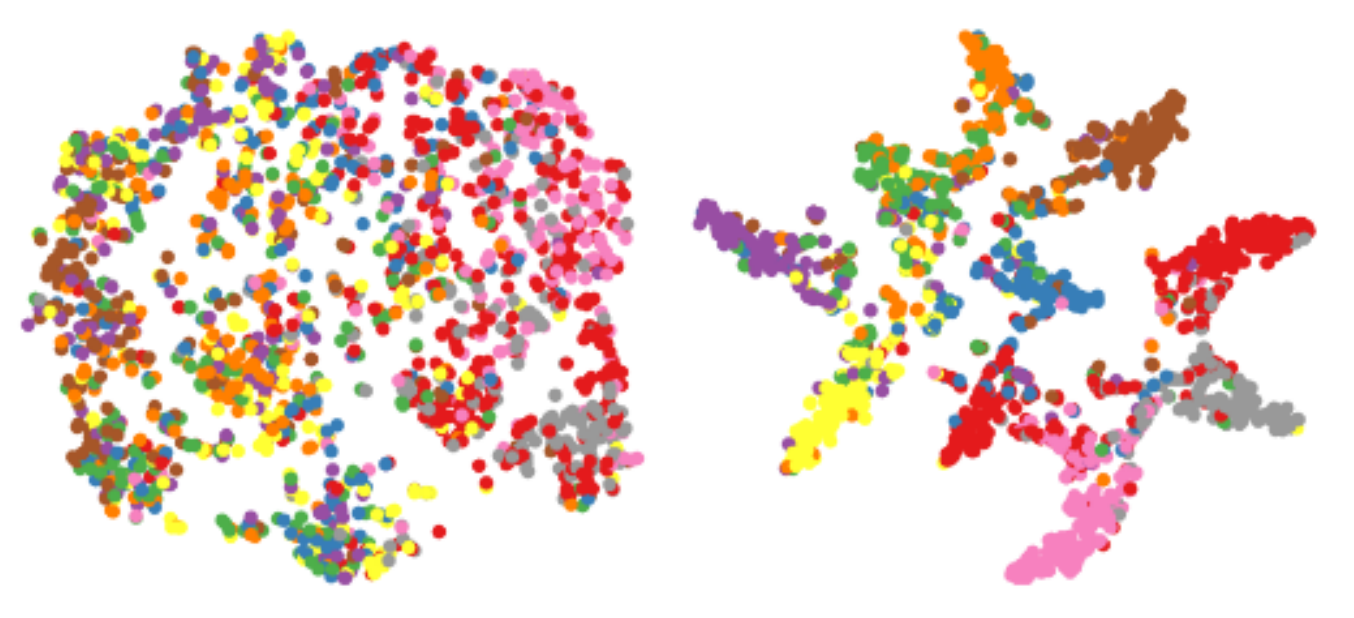}
\caption{Visualization of the samples in the learned feature space with the SPLIC method (right) and without defense (left).}
\label{fig:adv-clustering}
\end{figure}

The proposed SPLIC method is able to remove local image texture details, encourage the network to focus on more discriminative semantic features, and improve the robustness of the network to adversarial noise. We expect that, in the learned feature space, images from different classes will have much better separation or larger margins  since a small perturbation will not push the image sample across the decision boundary. To demonstrate this, we use the t-SNE method \cite{maaten2008visualizing} to visualize the learned features on the CIFAR-10 dataset. Figure \ref{fig:adv-clustering} (left) shows the  visualization of image features without the SPLIC method. The right figure shows the result for the SPLIC method. We can see that, with the SPLIC pre-processing and training, images from different classes are much better separated, indicating significantly improved robustness against adversarial attacks. 

\subsubsection{Performance on the clean images}
\begin{figure}
\centering
{\includegraphics[width=1\linewidth, height =0.5\linewidth]{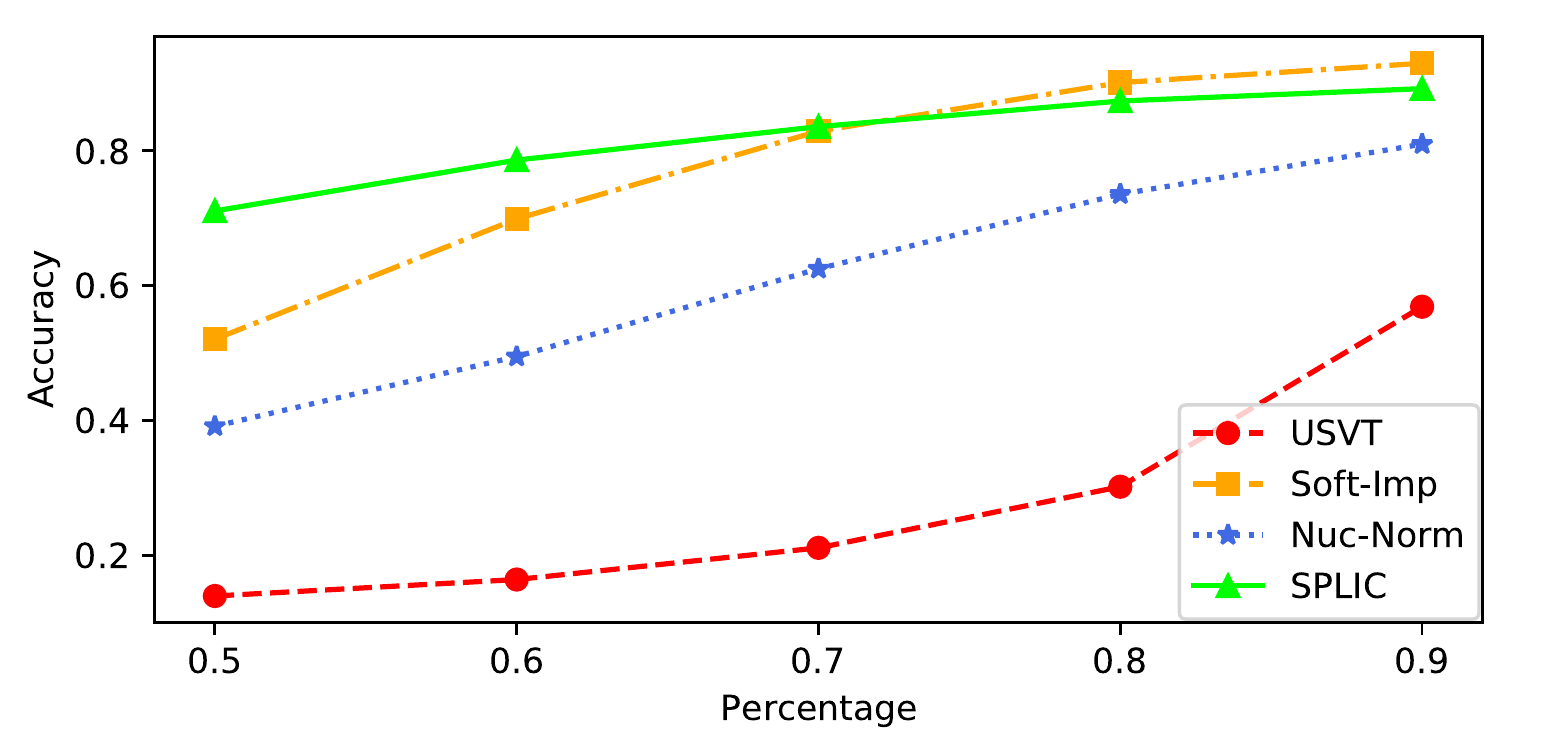}}
\caption{Reconstructions comparison on CIFAR-10 clean images with different probabilities of remained pixels.}
\label{fig:completion-compare-clean}
\end{figure}

In this part, we compare the performance of  our SPLIC algorithm on clean images with three low-rank approximation methods that have been implemented in \cite{yang2019me} with source code publicly available. These methods are Universal Singular Value Thresholding (USVT), Soft Impute (Soft-Imp) and Nuclear Norm (NUC-Norm). The the classification accuracy are shown in Figure \ref{fig:completion-compare-clean}. We test these reconstruction methods with different percentages of pixels used for anchor points. We can see  that, on the clean images without attack noise, our algorithm can maintain very good accuracy at different percentages of anchor points. This suggests that our method is able to preserve the important semantic structures in the original images.  

\subsubsection{Subjective examples of SPLIC results}

Figure \ref{fig:ori-adv-srf-splic-samples} shows seven examples of SPLIC processing results. The first row shows the original images from CIFAR-10. The second row shows the attacked images by the white-box PGD method. The attack noise is clearly visible. The third row shows the restored images by our SPLIC method. We can see  that the attack noise has been largely removed, the detailed image textures which are not important for recognition has been smoothed out, while important semantic structures are well preserved.

\subsubsection{Important of TV regularization in SPLIC}
\label{sec:sup:srf-compare}

\begin{figure}
\centering
\includegraphics[width=1\linewidth]{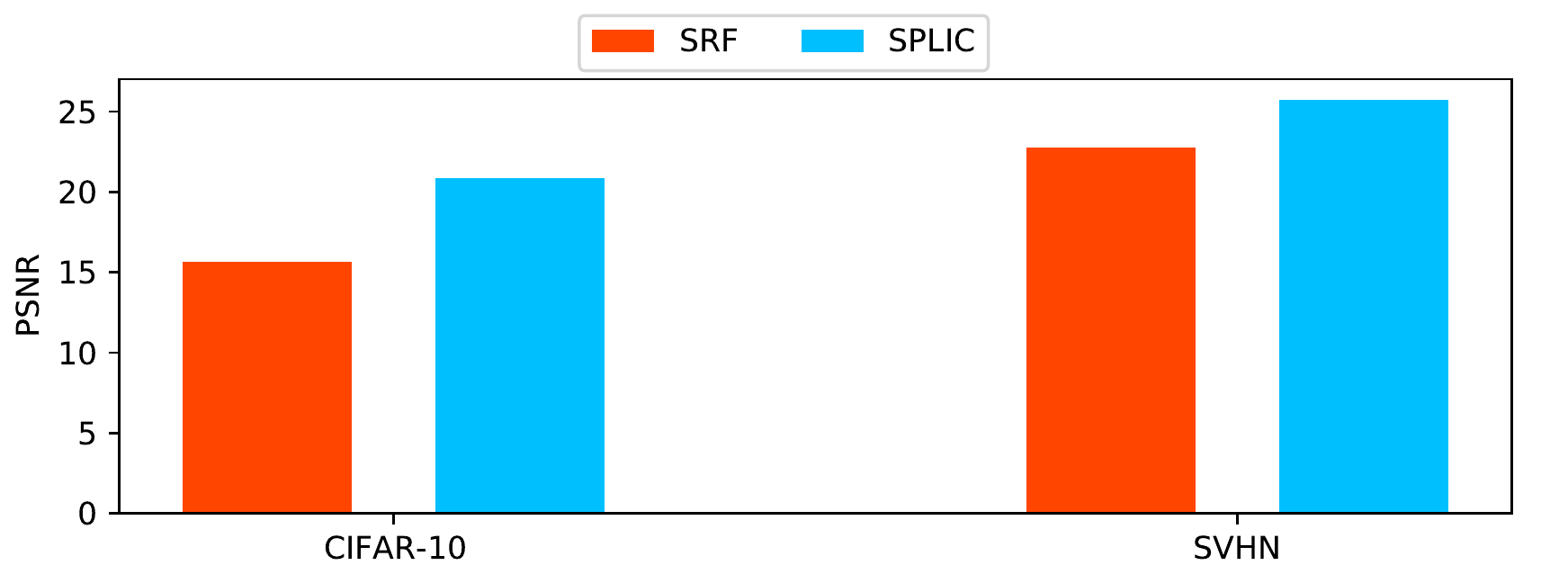}
\caption{PSNR of SRF and SPLIC.}
\label{fig:psnr-compare}
\end{figure}

In our proposed SPLIC method, we have incorporated the TV regularization into the smooth rank function. The remarkable performance achieved in our experiments are achieved by the unique combination of the progressive smooth rank function and the TV regularization. 
In the following experiments, we demonstrate that this TV constraint is very important. Figure \ref{fig:psnr-compare} shows the image quality obtained our SPLIC with (blue) and without the TV constraint (labeled with SRF in red). Here, we measure the image quality by the PSNR (peak signal to noise ratio) between the original image and the reconstructed image.
We can see that, by incorporating the TV constraint into the objective function, the image quality can be significantly improved. Table  \ref{tab:graybox-srf-splic-svhn} shows the gray-box defense performance comparison with SRF which is the SPLIC method without TV constraint. Since SVHN is a digit image dataset and these images structure is very simple, the SRF reconstruction on SVHN perform better than its reconstruction on CIFAR-10. We can see that our SPLIC method outperforms the SRF method on both clean image reconstruction and adversarial defending. It also demonstrates the importance of the TV constraint in our SPLIC objective function.

\begin{table}
\centering
\caption{Gray-box defense comparison without network training on SVHN.}
\label{tab:graybox-srf-splic-svhn}
\begin{tabular}{c|c|c|c|c}
\toprule
Method  &   Clean  &  FGSM  &  PGD  &  CW  \\
\hline
SRF &   $89.6\%$  &  $57.0\%$  &  $54.6\%$  &  $83.5\%$  \\
\hline
SPLIC &  $\textbf{95.9\%}$  &  $\textbf{59.8\%}$  &  $\textbf{63.6\%}$  &  $\textbf{86.0\%}$  \\
Gain & \textbf{+6.3\%} & \textbf{+2.8\%} & \textbf{+9.0\%} & \textbf{+2.5\%} \\
\bottomrule
\end{tabular}
\end{table}

\section{Conclusion}
\label{sec:conclusion}

In this work, we observed that the adversarial attack operates at local image textures as a special noise while the human visual system focuses on semantic structures and global visual cues. 
Motivated by this, we developed a structure-preserving progressive low-rank image completion (SPLIC) method to remove unneeded texture details from the input images and let the deep neural network focuses more on global object structures and semantic cues. We formulate the problem into a low-rank matrix completion problem with progressively smoothed rank functions to avoid local minimums and total variation constraint to enforce local smoothness during the optimization process. 
Our experimental results demonstrate that the proposed method is able to successfully remove the insignificant local image details and let the network learning focus on global object structures. On black-box, gray-box, and white-box attacks, our method outperforms existing defense methods and significantly improves the adversarial robustness of the network.

\printcredits

\bibliographystyle{cas-model2-names}

\bibliography{cas-refs}





\end{document}